\RequirePackage{fix-cm}
\documentclass{article}
\PassOptionsToPackage{nonamebreak}{natbib}
\usepackage{eviscene_fairmeta,times}

\usepackage{amsmath,amsfonts,bm}

\def\eqref#1{equation~\ref{#1}}

\def\1{\bm{1}}

\DeclareMathAlphabet{\mathsfit}{\encodingdefault}{\sfdefault}{m}{sl}
\SetMathAlphabet{\mathsfit}{bold}{\encodingdefault}{\sfdefault}{bx}{n}

\usepackage{hyperref}
\usepackage{url}
\usepackage{xurl}

\usepackage{graphicx}
\usepackage{flafter}
\usepackage{placeins}
\usepackage{needspace}
\usepackage{array}
\usepackage{booktabs}
\usepackage{colortbl}
\usepackage{tabularx}
\usepackage{longtable}
\usepackage{wrapfig}
\usepackage{listings}
\usepackage[most]{tcolorbox}
\usepackage{pifont}
\usepackage{tikz}
\usetikzlibrary{arrows.meta,fit,positioning,shapes.geometric}

\title{EviScene: Beyond ``Is It AI?''
for Agentic Image Factuality Assessment}

\author{%
  \makebox[\linewidth][c]{\begin{minipage}{\linewidth}
    \centering\normalfont
    {\fontsize{10}{13.5}\selectfont
      ZiAn Wang\textsuperscript{1,*},
      Ruiqi Liu\textsuperscript{2,*},
      Zhengbo Zhang\textsuperscript{2,*},
      Yi Han\textsuperscript{3},
      Jialiang Shen\textsuperscript{4},
      Ruoxin Chen\textsuperscript{5}\\
      Manni Cui\textsuperscript{6},
      Ximo Zhu\textsuperscript{2},
      Jin Jiang\textsuperscript{7},
      Mingzhe Liu\textsuperscript{8},
      Linhao Wang\textsuperscript{1},
      Wenhao Wang\textsuperscript{9,\ensuremath{\dagger}},
      Shu Wu\textsuperscript{2,\ensuremath{\dagger}}\endgraf}
    \vspace{5pt}
    {\fontsize{8.5}{11}\selectfont
      \textsuperscript{1}Jilin University\quad
      \textsuperscript{2}Institute of Automation, Chinese Academy of Sciences\quad
      \textsuperscript{3}Southwest University\\
      \textsuperscript{4}University of Sydney\quad
      \textsuperscript{5}Tencent Youtu Lab\quad
      \textsuperscript{6}Huazhong University of Science and Technology\\
      \textsuperscript{7}United Arab Emirates University\quad
      \textsuperscript{8}Shenzhen University\quad
      \textsuperscript{9}Vast Intelligence Lab\endgraf}
    \vspace{5pt}
    {\fontsize{8.5}{10.5}\selectfont
      \textsuperscript{*}Equal contribution.\qquad
      \textsuperscript{\ensuremath{\dagger}}Corresponding authors.\endgraf}
  \end{minipage}}%
}

\hypersetup{
  colorlinks=false,
  pdfborder={0 0 1},
  citebordercolor={0 1 0},
  linkbordercolor={1 0 0},
  urlbordercolor={0 1 1},
  pdftitle={EviScene: Beyond Is It AI? for Agentic Image Factuality Assessment},
  pdfauthor={ZiAn Wang, Ruiqi Liu, Zhengbo Zhang, Yi Han, Jialiang Shen, Ruoxin Chen, Manni Cui, Ximo Zhu, Jin Jiang, Mingzhe Liu, Linhao Wang, Wenhao Wang, Shu Wu}
}

\newcommand{\ours}{EviScene}
\newcommand{\task}{Image Factuality Assessment}
\newcommand{\benchmark}{EviLens}
\newcommand{\supported}{supported}
\newcommand{\refuted}{refuted}

\definecolor{tasknavy}{HTML}{18324B}
\definecolor{taskblue}{HTML}{E7F0F8}
\definecolor{taskblueaccent}{HTML}{4A78A8}
\definecolor{taskgreen}{HTML}{E4F3EA}
\definecolor{taskgreenaccent}{HTML}{2F7D50}
\definecolor{taskred}{HTML}{FBE8E7}
\definecolor{taskredaccent}{HTML}{B64A48}
\definecolor{taskgold}{HTML}{F8EFD8}
\definecolor{taskline}{HTML}{A9B8C5}
\definecolor{taskmuted}{HTML}{637383}
\definecolor{resultshade}{HTML}{EDF4F8}
\definecolor{resultgroup}{HTML}{F1F2F4}
\newcommand{\resulthead}[2]{%
  \multicolumn{1}{#1}{\(\vcenter{\hbox{\bfseries\shortstack[#1]{#2}}}\)}}
\newcommand{\yesmark}{\textcolor{taskgreenaccent}{\ding{51}}}
\newcommand{\nomark}{\textcolor{taskredaccent}{\ding{55}}}
\newtcolorbox{promptbox}[1]{
  enhanced jigsaw,
  breakable,
  colframe=black,
  colback=white,
  colbacktitle=black,
  coltitle=white,
  fonttitle=\bfseries\rmfamily,
  title={#1},
  boxrule=0.7pt,
  arc=1.8mm,
  outer arc=1.8mm,
  left=4mm,
  right=4mm,
  top=2.5mm,
  bottom=2.5mm,
  before skip=7pt,
  after skip=8pt,
  segmentation style={solid,black}
}
\newcommand{\promptfile}[3][\fontsize{7.5}{8.2}\selectfont]{%
  \par%
  \tcbinputlisting{
    enhanced jigsaw,
    breakable,
    lines before break=4,
    listing only,
    colframe=black!45,
    colback=white,
    colbacktitle=black!7,
    coltitle=black,
    fonttitle=\small\bfseries\rmfamily,
    title={#2},
    title after break={#2 (continued)},
    boxrule=0.5pt,
    arc=0.8mm,
    outer arc=0.8mm,
    left=2.5mm,
    right=2.5mm,
    top=1.4mm,
    bottom=1.4mm,
    before skip=6pt,
    after skip=6pt,
    listing file={#3},
    listing options={
      basicstyle=\normalfont\fontencoding{T1}\fontfamily{lmtt}\selectfont#1,
      breaklines=true,
      breakatwhitespace=false,
      columns=fullflexible,
      keepspaces=true,
      showstringspaces=false,
      tabsize=2
    }
  }%
}

\begin{document}

\maketitle
\raggedbottom

\begin{abstract}
Determining whether an image is AI-generated does not establish whether the
situation it depicts is factually supported. We formulate Image Factuality
Assessment: given an image and a generic verification instruction, a system
returns a supported or refuted verdict with an evidence-grounded explanation.
Without a supplied textual claim, the system must progressively refine factual
questions during investigation and align external evidence with the specific
situation depicted. We introduce EviLens, an expert-reviewed benchmark of
10,174 supported and refuted images from controlled construction and web
collection, covering twelve factual error types and evaluating both verdict
accuracy and evidence sufficiency. We develop EviScene, a multimodal ReAct agent
that interleaves visual inspection, external retrieval, and evidence analysis,
using new findings to refine its questions and revisit relevant image details.
We train a compact policy with supervised fine-tuning on reviewed teacher
trajectories, followed by privileged on-policy self-distillation applied to
verified decisions from the policy's own investigations. On the 1,527-image test set, the trained Qwen3.5-9B agent
achieves 79.05\% balanced accuracy and a Strict Evidence Sufficiency Rate
(SESR) of 52.26\%, improving over the same-backbone base agent by 6.71 and
9.37 percentage points, respectively. Its SESR is the highest among the
evaluated agents.
\par\smallskip
{\small Code, models, and data will be released at
\url{https://github.com/wangzian828/EVISCENE} and
\mbox{\url{https://huggingface.co/ZiAnwang/EviScene}}.}
\end{abstract}

\section{Introduction}

\begin{wrapfigure}[17]{r}{\dimexpr(\textwidth-\columnsep)/2\relax}
\centering
\vspace{-18pt}
\includegraphics[width=\linewidth]{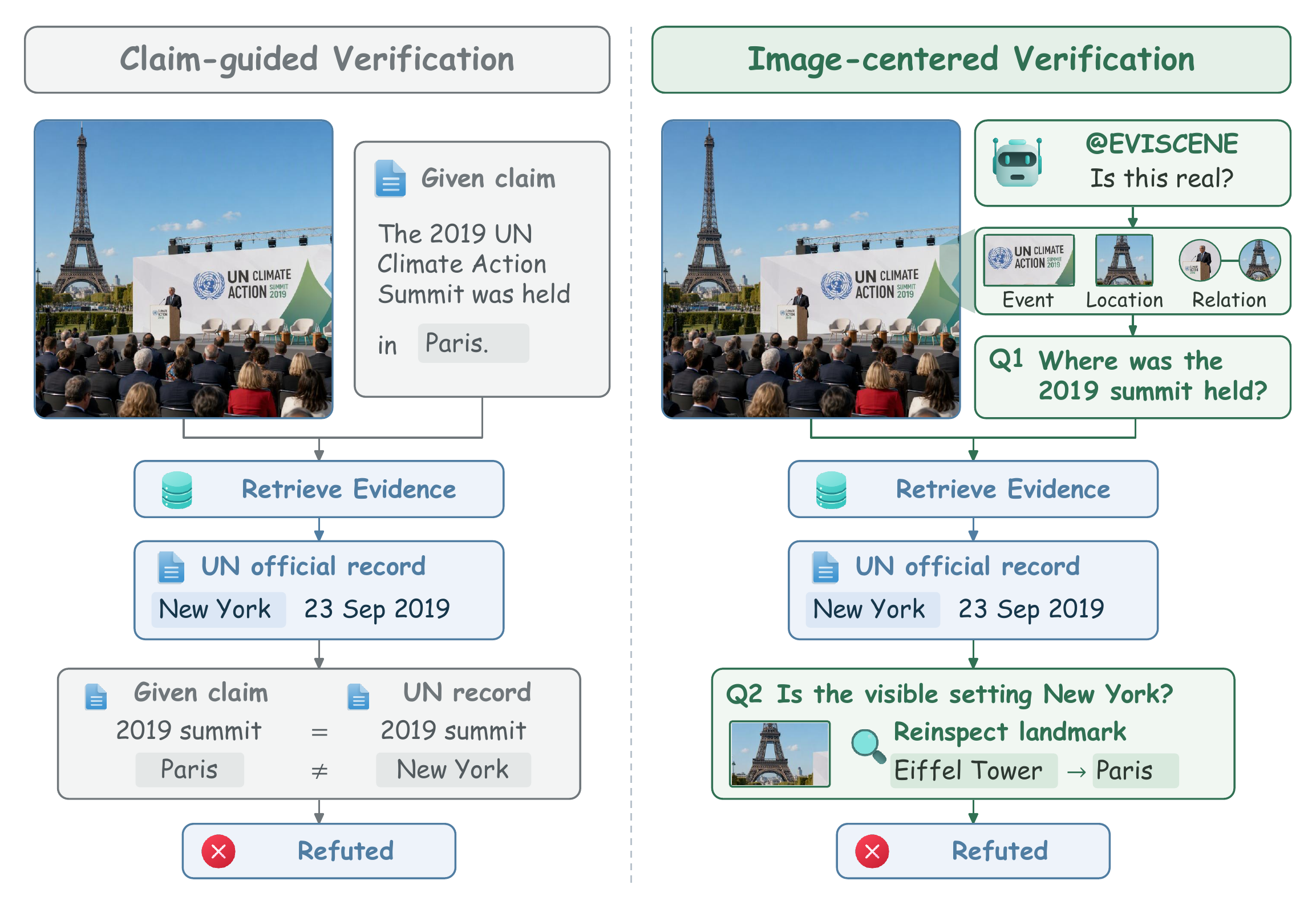}
\caption{Claim-guided versus image-centered verification. \ours{} alternates
evidence retrieval and visual reinspection to refine questions about the
depicted event and its context.}
\label{fig:task}
\end{wrapfigure}

Visual fact-checking uses external evidence to assess the factual accuracy of
image content and explain the resulting judgment. Recent work has explored
automating this process by combining multimodal models with web retrieval,
image search, and evidence analysis
\citep{akhtar2023multimodal,qi2024sniffer,pmlr-v267-braun25b}.
We focus on image-centered verification, asking whether the depicted situation
has factual support rather than how the image was produced.\looseness=-1

We refer to this task as \emph{\task}: given a single image and a generic
verification instruction, but no supplied textual claim, a system returns a
\supported{} or \refuted{} verdict with an evidence-grounded explanation. The
assessment concerns the factual situation conveyed by the image's visible
content, including the identities, activities, relationships, and
spatiotemporal context that determine its interpretation.\looseness=-1

This setting presents two main challenges. First, an initial visual inspection
is often insufficient to clarify and verify the complex factual situation
depicted in an image. In claim-centered benchmarks such as AVerImaTeC, a
supplied textual statement anchors evidence retrieval
\citep{NEURIPS2025_0695cbd1}. In our setting, the model instead draws on visual
content and newly obtained evidence to formulate and refine questions about
facts central to the image's meaning. Second, retrieved evidence must be
aligned with the specific situation depicted in the image. Material about the
same person or place may concern a different activity, year, or stage; a shared
person or place alone is insufficient to support the verdict. Addressing these
challenges requires interleaving visual inspection with evidence gathering,
using intermediate findings to refine what should be checked next.
Figure~\ref{fig:task} contrasts this image-centered setting with claim-guided
verification.

\begin{figure}[!t]
    \centering
    \includegraphics[width=\textwidth]{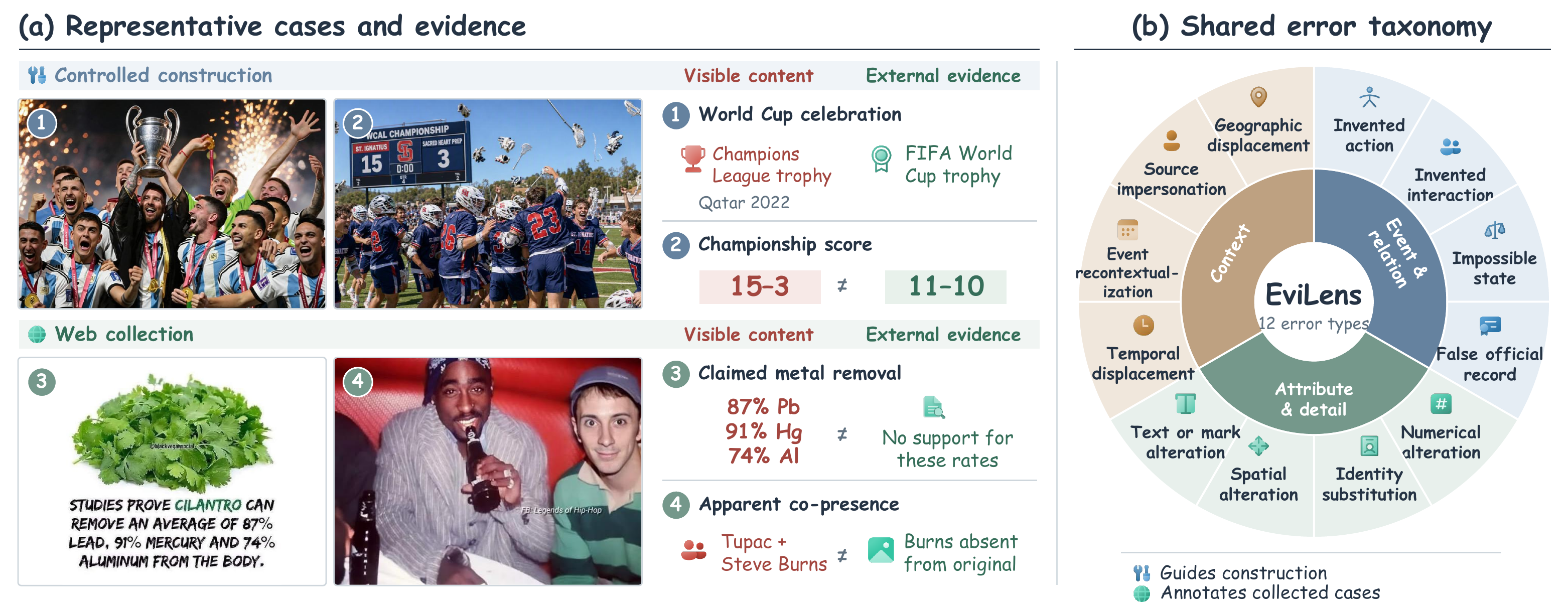}
    \caption{Factual errors in \benchmark{}: (a) constructed and collected
    examples with corresponding evidence; (b) the twelve error types grouped
    into three families.}
\label{fig:benchmark-taxonomy}
\end{figure}

To support research and evaluation in this setting, we introduce \benchmark,
a benchmark comprising 10,174 images: 3,114 supported and 7,060 refuted. Its
twelve factual error types fall into three families: event and relation,
attribute and detail, and context. We obtain images through controlled
construction and web collection. Controlled construction specifies and checks
the depicted factual content, including designated errors in refuted examples;
web collection preserves the diversity of naturally circulating content. All
examples undergo manual screening, and their factual descriptions are retained
together with the relevant construction or provenance records. Evaluation
considers both verdict accuracy and the sufficiency of the evidence presented
in the accompanying explanation.

We further develop \ours{}, a multimodal agent built on ReAct
\citep{yao2022react} that integrates visual inspection, web and image retrieval,
source reading, and reference-image comparison. At each step, the policy uses
the input image and accumulated investigation history to reason about an
unresolved factual question and select the next tool action. New observations
guide question refinement and subsequent actions, such as more targeted
searches or renewed inspection of relevant image regions. The final report
cites the observations supporting its verdict.

We train Qwen3.5-9B through supervised fine-tuning (SFT) on reviewed teacher
trajectories. We then apply privileged on-policy self-distillation (PSD;
\citealp{sleiman2026psd}) to selected decisions from the policy's own
investigations, using verified hint-guided continuations to supervise the
policy under its original context without exposing the hint to the student.\looseness=-1

On the 1,527-image \benchmark{} test set, \ours{} improves evidence sufficiency
over direct judgment across all five backbones evaluated in both settings. With
Qwen3.5-9B, the full SFT+PSD policy achieves 79.05\% balanced accuracy (BAcc)
and a Strict Evidence Sufficiency Rate (SESR) of 52.26\%. These scores exceed
those of the base agent with the same backbone and investigation interface by
6.71 and 9.37 percentage points, respectively. The full policy achieves the
highest SESR among all evaluated agents on the held-out test set.

Our contributions are threefold:
\begin{enumerate}
    \item We formulate \task{} and introduce \benchmark, a benchmark comprising
    10,174 supported and refuted images. It covers twelve factual error types
    and supports joint evaluation of verdict accuracy and evidence sufficiency.

    \item We develop \ours{}, a multimodal ReAct agent that interleaves visual
    analysis, external retrieval, and evidence inspection. It uses intermediate
    findings to refine factual questions and returns a verdict with an
    explanation citing the evidence used.

    \item We train a compact open-weight policy for multi-step visual
    fact-checking by applying SFT and PSD to reviewed trajectories and verified
    decisions. Across multiple backbones, we evaluate agent-based verification
    and analyze training effects, tool contributions, and performance across
    image sources and generators.\looseness=-1
\end{enumerate}

\section{Related Work}

\paragraph{Visual authenticity and factual verification.}
Visual authenticity research covers synthetic-content detection and factual
interpretation. One line of work focuses on determining whether an image or
video is AI-generated
(\citealp{NEURIPS2025_991c9f17}; \citeauthor{liu2025artifactsrealcentricenvelopemodeling}, \citeyear{liu2025artifactsrealcentricenvelopemodeling}, \citeyear{liu2026mirrormanifoldidealreference}; \citealp{cui2026rethinkingreadoutunlockingvideo}). Other work
evaluates whether depicted content is consistent with factual knowledge or commonsense
\citep{lim2025evaluating,huang-etal-2025-t2i,niu2026wise,10377616}.
Related work also studies factually grounded refinement and reasoning about
semantic visual anomalies in generated images
\citep{lim2026fagerfactuallygroundedevaluation,ICLR2026_b5b4d923}. Contextual verification examines
mismatches between images and their accompanying context and seeks to recover
an image's source, time, and location
\citep{luo-etal-2021-newsclippings,aneja2023cosmos,papadopoulos2024verite,tonglet-etal-2024-image}.
Evidence-based multimodal fact-checking verifies supplied claims using external
information
\citep{10.1145/3539618.3591879,NEURIPS2025_0695cbd1,qi2024sniffer}.
DEFAME \citep{pmlr-v267-braun25b} combines adaptive tool selection and iterative
evidence collection within a six-stage verification pipeline.
In \task{}, the system receives an image and a generic verification
instruction, without a supplied textual claim. \ours{} uses a single ReAct
policy to refine image-grounded questions and select each next tool action
from the accumulated interaction history, then returns an evidence-grounded
verdict. Its label reflects the
factuality of the depicted situation, regardless of how the image was produced.\looseness=-1

\paragraph{Multimodal agents and policy learning.}
ReAct interleaves reasoning with tool use to support multi-step investigation
\citep{yao2022react}. Multimodal search and browsing approaches use intermediate
findings to guide subsequent retrieval and visual analysis
\citep{zhang2026visualseekervisualnativemultimodalagentic,zhang2026visbrowsebenchbenchmarkingvisualnativesearch}. For policy
learning, agent fine-tuning uses interaction trajectories to supervise reasoning
and tool use \citep{chen2023fireactlanguageagentfinetuning,zeng-etal-2024-agenttuning}. Complementary
approaches adapt training data or supervision to states visited by the current
policy through on-policy data evolution \citep{huang2026onpolicydataevolutionvisualnative} and distillation
\citep{ICLR2024_5be69a58,sleiman2026psd}. \ours{} builds on this work for
image-centered verification, using new evidence to refine factual questions
and guide the reinspection of relevant image details. We first fine-tune the
policy on reviewed teacher trajectories and then apply privileged on-policy
self-distillation to decisions reached during its investigations.

\section{\benchmark}
\label{sec:benchmark}

\benchmark{} evaluates models' ability to verify the factual situation
conveyed by an image and justify their verdict with external evidence. The
benchmark contains 10,174 images from controlled construction and web
collection: 3,114 supported and 7,060 refuted.

\subsection{Twelve Factual Error Types}
\label{sec:dataset-taxonomy}

The taxonomy organizes factual discrepancies around three questions:
\emph{Did the depicted event occur, or did the depicted relation hold? Is the
visible attribute or detail correct? Is the content presented in its true
context?} The twelve error types are grouped into three corresponding families:
event and relation, attribute and detail, and context
(Figure~\ref{fig:benchmark-taxonomy}(b)). Figure~\ref{fig:benchmark-taxonomy}(a)
illustrates representative discrepancies by comparing visible content with the
corresponding external evidence.

\begin{figure}[!t]
    \centering
    \includegraphics[width=\textwidth]{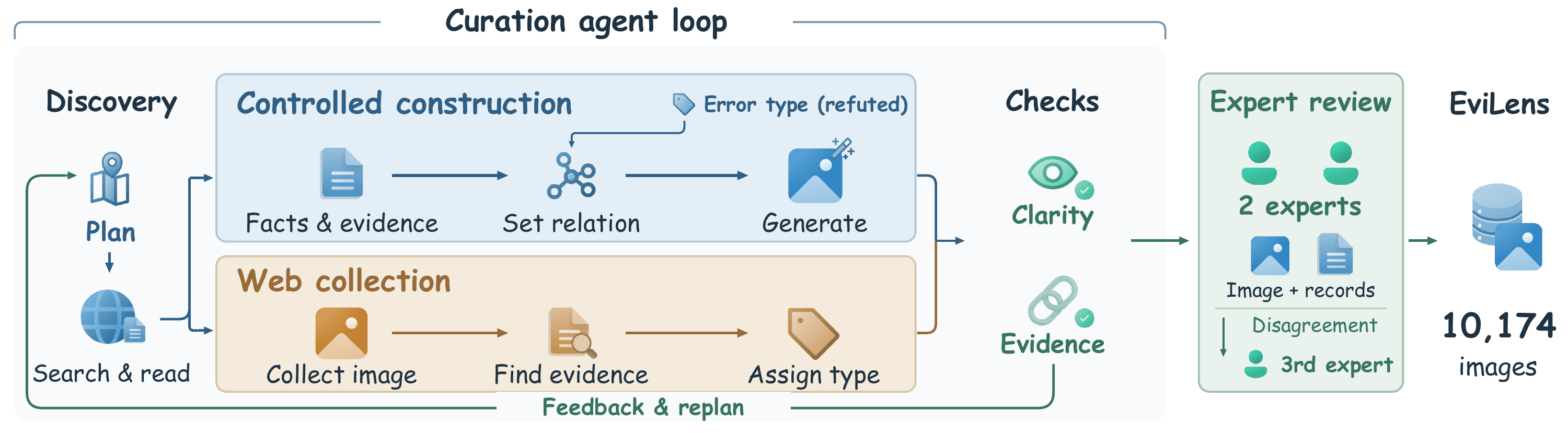}
    \caption{Construction and collection of \benchmark{}. The curation agent
    builds and checks candidates, uses feedback to guide later rounds, and
    submits them for expert review.}
    \label{fig:benchmark-pipeline}
    \vspace{-10pt}
\end{figure}

For each generated refuted image, we assign a primary error type to guide
construction and check whether the resulting image exhibits the intended
discrepancy. For collected refuted images, we assign error types after
acquisition to describe the factual error mechanism in each image. Operational
definitions and distinctions between neighboring types are given in
Appendix~\ref{app:taxonomy-definitions}. An image may contain several visible
details, but its primary type identifies the discrepancy that decides the
refuted label. This keeps the error analysis tied to the fact being verified.
The same taxonomy also tracks factual errors across constructed and collected
refuted images. Its category describes the error, not the image's acquisition
route, production method, or visual style.

\subsection{Construction and Collection}
\label{sec:dataset-acquisition}

The curation agent in Figure~\ref{fig:benchmark-pipeline} plans searches based
on coverage gaps, source diversity, and time windows, then reads the retrieved
sources and prepares candidate examples. For controlled construction, the agent
uses factual material to specify the content to be rendered and checks the
resulting image against that specification. For web collection, it starts from
existing images and recovers the factual descriptions and evidence associated
with their visible content, while retaining the diversity of naturally
circulating material. Both pathways retain evidence for their supported examples.\looseness=-1

Generated supported images depict situations established by factual references.
For generated refuted images, counterfactual construction alters a selected
detail or relation in a documented situation, whereas fabrication depicts an
occurrence involving real entities that conflicts with relevant records or
constraints. Web collection contributes supported images corroborated by
source material and refuted images from fact-checking sources. The collected
material includes both photographs and circulating synthetic images.

Automated checks assess visual clarity and evidence alignment and screen for
duplication and leakage. The results inform subsequent planning. Two domain
experts review each candidate image with its factual description and retained
records, checking visual clarity, the grounding of the verification target,
and the evidence supporting its interpretation and label. For constructed refuted images, they also
confirm that the intended discrepancy is visible. A third domain expert
resolves disagreements. Review outcomes are acceptance, acceptance with edits,
or rejection.\looseness=-1

Each retained example includes an image, its factual label and explanation,
relevant evidence, and the associated construction or provenance records.
The dataset is divided into 8,647 training images and 1,527 test images.
Related images from the same source or event are grouped when constructing
the splits. At evaluation, the task input consists only of the image and a
generic verification instruction. We use private reference material to check
the verdict and the grounds stated in each response. Appendix~\ref{app:construction}
details the five acquisition pathways, screening criteria, dataset composition,
and split construction, including source-wise label counts.

\begin{figure}[!t]
\centering
\includegraphics[width=0.98\textwidth]{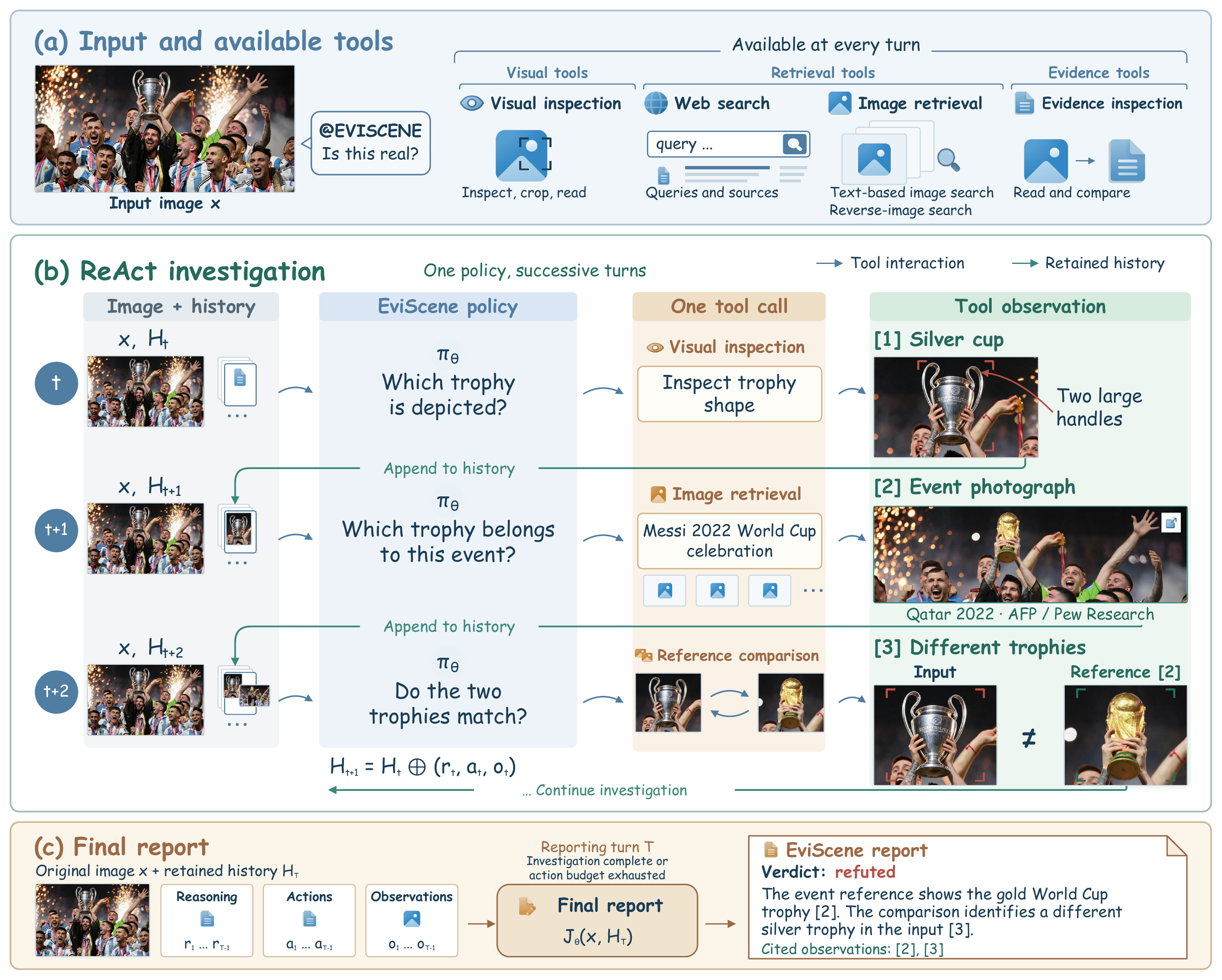}
\caption{\ours{} investigation and reporting. (a) Three tool groups available
at every investigation turn. (b) An illustrative ReAct sequence: observations
guide question refinement and tool selection, while each interaction is
appended to the history. (c) At reporting turn \(T\), the policy uses the input
image and full history of turns \(1,\ldots,T-1\) to generate a verdict,
explanation, and cited observations.}
\label{fig:pipeline}
\end{figure}

\section{\ours}

\ours{} combines an image-centered investigation agent with two-stage policy training. The agent interleaves visual inspection with external evidence gathering within a shared ReAct history \citep{yao2022react}. We train its policy through full-trajectory supervised fine-tuning followed by PSD on selected decisions from its own investigations. Figure~\ref{fig:pipeline} illustrates the investigation process; Appendix~\ref{app:implementation} gives the tool and training details.

\subsection{Image-Centered Multimodal Investigation}

Given an input image \(x\), we write the investigation history before turn \(t\) as
\[
H_t=(r_1,a_1,o_1,\ldots,r_{t-1},a_{t-1},o_{t-1}),
\]
where \(r_i\) is the policy's reasoning, \(a_i\) its tool action, and \(o_i\) the raw observation.

At each investigation turn \(t\), the policy \(\pi_\theta\) produces a short reasoning step and one native tool action:
\[
(r_t,a_t)\sim\pi_\theta(\cdot\mid x,H_t),\qquad
o_t=\mathcal{T}_{a_t}(x,H_t),
\]
where \(\mathcal{T}_{a_t}\) denotes the execution of tool action \(a_t\), which returns observation \(o_t\). We append the resulting tuple \((r_t,a_t,o_t)\) to \(H_t\) to obtain \(H_{t+1}\). This process repeats until the policy completes the investigation \mbox{or exhausts its action budget.}

\ours{} provides three tool groups. Visual tools inspect scenes and selected regions, read and localize text, count objects, examine relations, and perform evidence-guided visual reinspection. Retrieval tools support web search, text-based image search, and reverse-image search. Evidence tools inspect source pages and compare the input image with retrieved reference images.

Retrieved material about the same person or place may concern a different event, time, or stage. The policy therefore reads source pages or compares retrieved images with the input to determine whether the information supports or contradicts the fact being checked. The policy uses each observation to refine the current factual question and guide its next action. For example, an external identification can prompt it to reinspect the relevant image region, text, or relation.

Let \(T\) denote the final reporting turn, so that \(H_T\) contains all preceding tool interactions. The policy generates the final report from the input image and this history:
\[
(\hat{y},\hat{e},C)=J_\theta(x,H_T),
\]
where \(\hat{y}\in\{\mathrm{supported},\mathrm{refuted}\}\) is the factual verdict, \(\hat{e}\) is the explanation, and \(C\) contains the observations from \(H_T\) cited in the explanation as substantive evidence for the verdict.

\subsection{Full-Trajectory Supervised Fine-Tuning}

We use Qwen3.5-397B-A17B \citep{qwen3.5} to generate teacher trajectories through the same agent interface used at inference. Each trajectory records the model-visible reasoning, tool calls, observations, retrieved images, and final report. We retain a trajectory only if it is structurally complete and its \mbox{recorded observations support the final report.}

We perform full-parameter SFT on Qwen3.5-9B \citep{qwen3.5} using 4,872 reviewed teacher trajectories. Training runs for three epochs with a learning rate of \(10^{-5}\). Each training trajectory retains the context for every action and the final report. For a trajectory token sequence \(z_{1:n}\), we minimize the weighted loss
\[
\mathcal{L}_{\mathrm{SFT}}
=
-\sum_{i=1}^{n}w_i\log p_\theta(z_i\mid z_{<i}),
\]
where \(w_i=2\) for tool-call and final-report tokens, \(w_i=1\) for reasoning tokens, and \(w_i=0\) for instruction and tool-observation tokens. Instruction and tool-observation tokens remain in the context but do not contribute to the loss; associated images are also retained as conditioning inputs. This weighting emphasizes tool calls and the final report while retaining supervision on reasoning.

\begin{table}[!t]
\caption{Results (\%) on 1,527 test images. Best and second-best results
among \ours{} agents are bold and underlined, respectively; trained Qwen
policies are shaded. Full denotes the three-epoch SFT policy after PSD.}
\label{tab:main-results}
\centering
\small
\setlength{\tabcolsep}{3pt}
\renewcommand{\arraystretch}{1.14}
\begin{tabularx}{\textwidth}{@{}>{\raggedright\arraybackslash}p{0.27\textwidth}*{5}{>{\centering\arraybackslash}X}@{}}
\toprule
\resulthead{l}{Method} &
\resulthead{c}{BAcc} &
\resulthead{c}{Macro\\F1} &
\resulthead{c}{Supported\\recall} &
\resulthead{c}{Refuted\\recall} &
\resulthead{c}{SESR} \\
\midrule
\rowcolor{resultgroup}
\multicolumn{6}{l}{\textbf{Direct visual judgment}} \\
GPT-5.5 & 75.87 & 70.51 & 79.58 & 72.17 & 3.67 \\
GPT-5.2 & 65.84 & 55.41 & 85.94 & 45.74 & 2.95 \\
Gemini 3.7 Flash & 75.64 & 72.80 & 71.62 & 79.65 & 9.10 \\
Gemini 3.1 Pro & 62.58 & 64.54 & 28.65 & 96.52 & 5.57 \\
Claude Opus 4.8 & 72.18 & 68.67 & 69.76 & 74.61 & 6.75 \\
Kimi K2.6 & 70.24 & 62.44 & 81.17 & 59.30 & 8.91 \\
Doubao Seed 2.0 Pro & 70.16 & 63.77 & 77.45 & 62.87 & 8.45 \\
MiniMax-M3 & 66.66 & 56.56 & 84.62 & 48.70 & 3.41 \\
GLM-4.6V & 59.96 & 41.86 & 95.49 & 24.43 & 3.01 \\
Qwen3.5-397B-A17B & 68.68 & 67.82 & 56.23 & 81.13 & 6.68 \\
Qwen3.5-9B & 65.70 & 57.26 & 79.05 & 52.35 & 1.70 \\
\midrule
\rowcolor{resultgroup}
\multicolumn{6}{l}{\textbf{\ours{} agents}} \\
GPT-5.5 & 78.13 & 72.06 & \underline{83.82} & 72.43 & \underline{50.49} \\
Gemini 3.7 Flash & \textbf{81.57} & \textbf{76.70} & \textbf{84.35} & 78.78 & 42.89 \\
Gemini 3.1 Pro & 62.42 & 64.17 & 29.71 & \textbf{95.13} & 32.68 \\
Qwen3.5-397B-A17B & 71.72 & 73.76 & 50.66 & 92.78 & 43.29 \\
Qwen3.5-9B, Base & 72.34 & 74.57 & 51.19 & \underline{93.48} & 42.89 \\
\rowcolor{resultshade}
Qwen3.5-9B, SFT & 78.81 & \underline{75.69} & 76.39 & 81.22 & \underline{50.49} \\
\rowcolor{resultshade}
Qwen3.5-9B, Full & \underline{79.05} & 74.67 & 80.11 & 78.00 & \textbf{52.26} \\
\bottomrule
\end{tabularx}
\end{table}

\subsection{Privileged On-Policy Self-Distillation}

After SFT, we apply privileged on-policy self-distillation (PSD; \citealp{sleiman2026psd}) to selected decisions in \mbox{the policy's own investigations.}

\paragraph{Decision localization and continuation verification.}
We use Gemini 3.7 Flash \citep{google2026gemini37flash} as a training-only intervention model to localize decisions, construct hints, and review continuations. At the start of each PSD round, we freeze a copy of the current policy as \(\pi_{\bar{\theta}}\) and collect new investigation trajectories. From these trajectories, the localizer selects a student decision at turn \(t\) involving search direction, event association, visual evidence, or the interpretation of retrieved evidence. It is instructed to identify the earliest recoverable error and support its selection with the original step index and recorded observations, rather than relying solely on an incorrect final verdict.\looseness=-1

The hint constructor may consult training-only information to propose a short procedural hint \(h_t\). At the selected decision, the hint provides procedural guidance without revealing the answer, specifying an exact tool action or query, or supplying a case-specific solution. We insert \(h_t\) into the original context \(c_t\) at the selected decision to form the teacher context \(\tilde{c}_t\):
\[
\tilde{c}_t=\operatorname{insert}(c_t,h_t).
\]

The frozen Qwen policy \(\pi_{\bar{\theta}}\) continues the investigation from \(\tilde{c}_t\) until it generates the final report. We accept the continuation for training only if the selected decision is corrected, the final report is correct and evidence-supported, and the trajectory passes the audit. The complete continuation is used for verification, whereas distillation uses the frozen policy's token-level probability distributions only for the selected assistant completion.

\paragraph{Privileged self-distillation.}
The student is trained to match the frozen teacher's top-\(K\) token probabilities at each position in the selected assistant completion. We also extract preservation targets from successful investigations to retain behavior supported by complete investigations.

Let \(\mathcal{D}_{\mathrm{PSD}}\) denote the set of accepted repair and preservation targets. For each target \(d\in\mathcal{D}_{\mathrm{PSD}}\), \(u_d\) denotes the token sequence of the selected assistant completion, and \(c_d^T\) and \(c_d^S\) denote the teacher and student contexts. For a repair target associated with turn \(t\), \(c_d^T=\tilde{c}_t\) and \(c_d^S=c_t\); for a preservation target, \(c_d^T=c_d^S\). Each target retains the recorded image and tool context.

At each position \(\ell\) in \(u_d\), the frozen teacher distribution is conditioned on \(c_d^T\) and the completion prefix \(u_{d,<\ell}\). We retain its top-\(K\) tokens \(\mathcal{K}_{d,\ell}\) and renormalize their probabilities to obtain \(\bar{q}_{d,\ell}\). We minimize the distillation loss
\[
\mathcal{L}_{\mathrm{PSD}}
=
-\sum_{d\in\mathcal{D}_{\mathrm{PSD}}}
\sum_{\ell=1}^{|u_d|}
\sum_{v\in\mathcal{K}_{d,\ell}}
\bar{q}_{d,\ell}(v)
\log p_\theta\!\left(v\mid c_d^S,u_{d,<\ell}\right).
\]

Both target types receive the same per-target weight, and the loss is summed over the supervised token positions in each selected completion. The updated policy then generates new investigation trajectories for the next round.

\begin{figure}[!t]
\centering
\includegraphics[width=\textwidth]{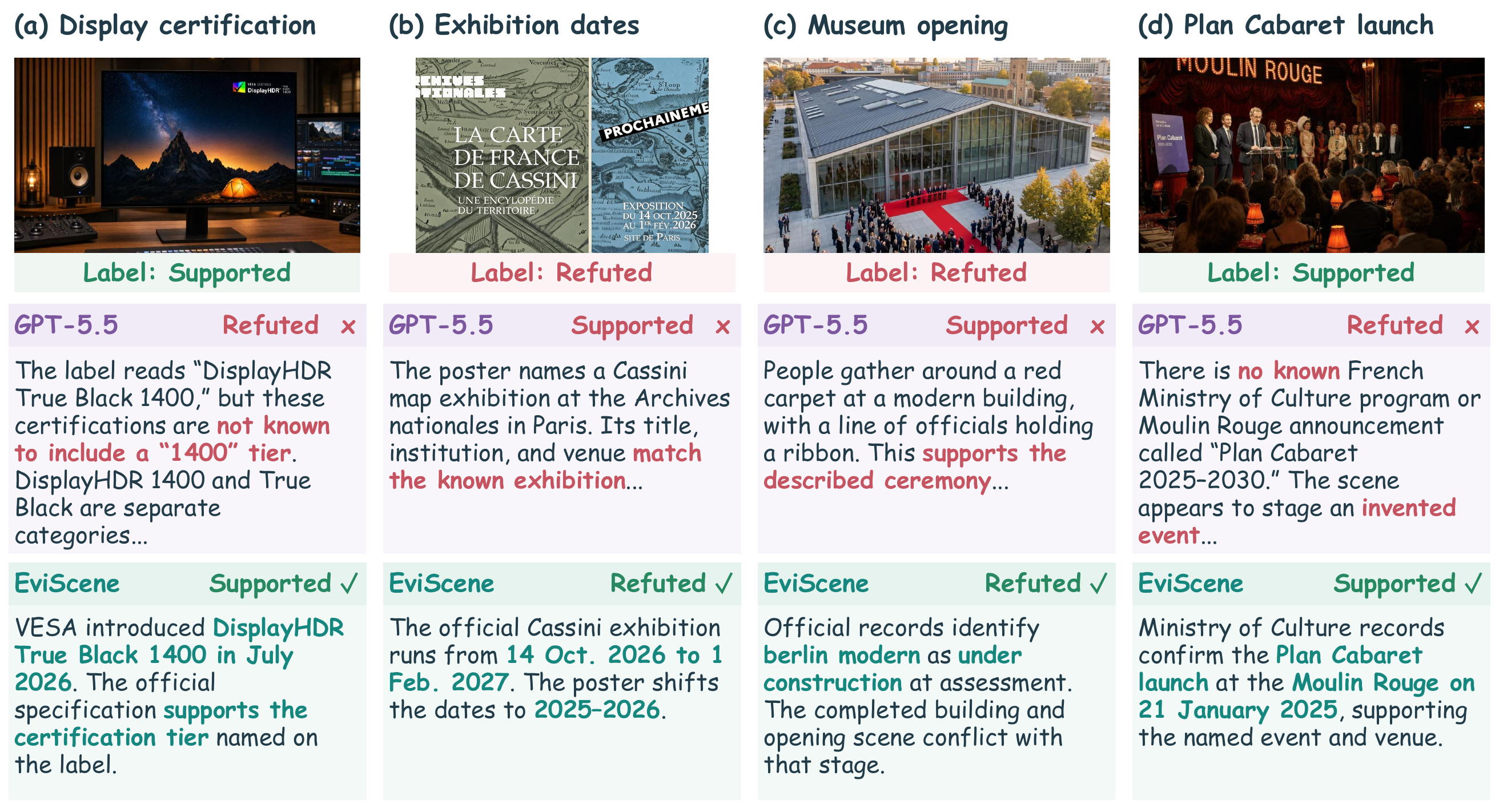}
\caption{Qualitative examples from \benchmark{}. Each column pairs an image
and its label with the GPT-5.5 and \ours{} verdicts and the decisive factual
comparison.}
\label{fig:qualitative-main}
\end{figure}

\Needspace{4\baselineskip}
\section{Experiments}
\label{sec:experiments}

\subsection{Experimental Setup and Evaluation Protocol}

\paragraph{Test set.}
The \benchmark{} test set contains 1,527 images: 377 supported and 1,150 refuted.

\paragraph{Compared systems.}
We compare direct judgment and \ours{} on five shared backbones: GPT-5.5,
Gemini~3.7 Flash, Gemini~3.1 Pro, Qwen3.5-397B-A17B, and Qwen3.5-9B.
Additional direct-judgment baselines include GPT-5.2, Claude Opus~4.8,
Kimi K2.6, Doubao Seed 2.0 Pro, MiniMax-M3, and GLM-4.6V
(Table~\ref{tab:main-results}). For Qwen3.5-9B, we evaluate direct judgment
(Direct), the base agent (Base), the SFT policy (SFT), and the SFT+PSD policy
(Full). All agents share a runtime with a maximum of 24 actions per
investigation. Web and text-based image search use Serper, and reverse-image
search uses Serper Lens. Jina Reader retrieves webpage content, while the
Baidu general OCR API transcribes and localizes visible text. Fact-checking
websites are blocked from evidence access during investigation. Model-mediated
visual inspection and page-evidence extraction
use separate requests to the agent's own backbone.
For Qwen3.5-9B, these requests use the checkpoint corresponding to the
evaluated configuration: Base, SFT, or Full.
Appendix~\ref{app:implementation} gives the model sources and tool configurations.
For each shared backbone, Direct answers from the image, while \ours{} can
inspect tool observations before its final judgment. The additional direct
baselines provide reference points for the same image-level task.
Both settings use the same held-out test images and final-response criteria.

\paragraph{Metrics.}
We report balanced accuracy (BAcc), macro F1, and class-wise recall. BAcc
averages supported and refuted recall. Strict Evidence Sufficiency Rate (SESR)
is the percentage of cases with a correct verdict and sufficient grounds for
the decisive image-grounded fact.

\paragraph{Evaluation protocol.}
We use Gemini~3.7 Flash \citep{google2026gemini37flash} to assess each final
response against private reference material. The shared criteria cover verdict
correctness, factual alignment, and the sufficiency of the grounds stated in
the response. Appendix~\ref{app:evaluation} provides the full metric
definitions and evaluation criteria.

\begin{table}[!t]
\caption{Results (\%) by image source and generator for Qwen3.5-9B. Each entry
reports BAcc / SESR; generator columns cover controlled-construction images.
Best and second-best values for each column and metric are bold and underlined.}
\label{tab:source-generator-results}
\centering
\small
\setlength{\tabcolsep}{3pt}
\renewcommand{\arraystretch}{1.50}
\begin{tabularx}{\textwidth}{@{}l*{5}{>{\centering\arraybackslash}X}@{}}
\toprule
& \multicolumn{2}{c}{\textbf{Acquisition source}} &
  \multicolumn{3}{c}{\textbf{Controlled test generator}} \\
\cmidrule(lr){2-3}\cmidrule(lr){4-6}
\textbf{Policy} &
\resulthead{c}{Controlled} &
\resulthead{c}{Web} &
\resulthead{c}{Nano Banana\\Pro} &
\resulthead{c}{GPT Image 2} &
\resulthead{c}{Seedream 5.0\\Lite} \\
\midrule
Direct & 61.11 / 1.29 & 74.46 / 3.46 & 62.54 / 1.17 & 52.95 / 0.71 & 49.39 / 2.32 \\
Base agent & 67.87 / 43.94 & 73.34 / 38.41 & 69.43 / 32.36 & 66.36 / 55.56 & 60.20 / 47.35 \\
SFT & \underline{76.39} / \underline{50.97} & \underline{76.17} / \underline{48.44} & \textbf{77.36} / \underline{39.18} & \underline{73.74} / \textbf{61.47} & \underline{67.63} / \textbf{56.29} \\
\rowcolor{resultshade}
Full & \textbf{76.75} / \textbf{52.75} & \textbf{77.21} / \textbf{50.17} & \underline{75.87} / \textbf{46.98} & \textbf{78.75} / \underline{58.39} & \textbf{69.68} / \underline{54.64} \\
\bottomrule
\end{tabularx}
\end{table}

\subsection{Main Results}

\paragraph{Investigation improves evidence sufficiency across backbones.}
Across the five shared backbones in Table~\ref{tab:main-results}, \ours{}
improves SESR over direct judgment by 27.11 to 46.82 percentage points. BAcc
also improves for four of the five backbones. For Qwen3.5-9B, the Base agent
raises BAcc from 65.70\% to 72.34\% and SESR from 1.70\% to 42.89\%, without
SFT or PSD. Some direct models show pronounced verdict preferences: Gemini~3.1
Pro recalls 96.52\% of refuted but only 28.65\% of supported cases, while
GLM-4.6V has 95.49\% supported and 24.43\% refuted recall.
The exception to the BAcc trend is Gemini~3.1 Pro: its BAcc changes from
62.58\% to 62.42\% with investigation, even as SESR rises from 5.57\% to
32.68\%.
For the other three paired backbones, investigation changes BAcc by 2.26
points for GPT-5.5, 5.93 for Gemini~3.7 Flash, and 3.04 for
Qwen3.5-397B-A17B. Their SESR gains are 46.82, 33.79, and 36.61 points,
respectively.
Gemini~3.7 Flash gains 12.73 points in supported recall and loses 0.87 in
refuted recall; Qwen3.5-397B-A17B loses 5.57 and gains 11.65 points,
respectively.

\begin{wraptable}[7]{R}{0.49\textwidth}
\centering
\small
\caption{Tool-family ablations (\%).}
\label{tab:ablations}
\setlength{\tabcolsep}{4pt}
\renewcommand{\arraystretch}{1.05}
\begin{tabular*}{\linewidth}{@{\extracolsep{\fill}}lrr@{}}
\toprule
\textbf{Variant} & \textbf{BAcc} & \textbf{SESR} \\
\midrule
All tools (Full) & \textbf{79.05} & \textbf{52.26} \\
w/o web search & 74.37 & 35.89 \\
w/o image retrieval & 74.52 & 40.01 \\
w/o evidence inspection & 76.55 & 46.04 \\
\bottomrule
\end{tabular*}
\end{wraptable}

\paragraph{The full compact policy leads in SESR.}
The full Qwen3.5-9B policy achieves the highest SESR (52.26\%) and
second-highest BAcc (79.05\%) among the evaluated agents. It also outperforms
the substantially larger Qwen3.5-397B-A17B agent by 7.33 percentage points in
BAcc and 8.97 percentage points in SESR under the same action limit.

\subsection{Training and Tool Analysis}
\label{sec:agent-ablations}

\paragraph{Training-stage analysis.}
For Qwen3.5-9B, Base recalls 93.48\% of refuted cases but only 51.19\% of
supported cases. SFT raises supported recall to 76.39\% while refuted recall
falls to 81.22\%, moving BAcc from 72.34\% to 78.81\% and SESR from 42.89\%
to 50.49\%. SFT narrows the gap between supported and refuted recall from
42.29 to 4.83 percentage points. After PSD, supported recall
rises to 80.11\% and refuted recall falls to 78.00\%; their gap narrows from
4.83 to 2.11 percentage points. Relative to SFT, Full gains 0.24 points in
BAcc and 1.77 points in SESR. Macro F1 is 75.69\% for SFT and 74.67\% for
Full. Overall, Full exceeds Base by 6.71 BAcc points and 9.37 SESR points.\looseness=-1

\paragraph{Tool ablations.}
We disable one external-evidence tool family at a time from the full
Qwen3.5-9B policy (Table~\ref{tab:ablations}), retaining direct visual
operations. All variants use the same prompt, action budget, judgment
procedure, and test set.\looseness=-1

Relative to Full, removing web search lowers BAcc/SESR by 4.68/16.37 points;
image retrieval by 4.53/12.25; and evidence inspection by 2.50/6.22.
SESR falls more in every case. Appendix~\ref{app:agent-configuration} lists
the exact operations disabled for each tool family.

\subsection{Source and Generator Analysis}

\paragraph{Acquisition sources.}
Table~\ref{tab:source-generator-results} reports results by image source and
generator. From Direct to Full, SESR rises from 1.29\% to 52.75\% on constructed
images and from 3.46\% to 50.17\% on web-collected images. Relative to Base,
Full improves SESR by 8.81 and 11.76 percentage points on the two sources,
respectively. Full also achieves similar BAcc on both, at 76.75\% and 77.21\%.

\paragraph{Image generators.}
Full improves both BAcc and SESR over Base on images from all three generators.
It achieves 46.98\% SESR on Nano Banana Pro, 58.39\% on GPT Image~2, and 54.64\%
on Seedream~5.0 Lite. GPT Image~2 and Seedream~5.0 Lite are not used to construct
training images, so the gains over Base are not limited to the training
generator.\looseness=-1

\subsection{Qualitative Analysis}

Figure~\ref{fig:qualitative-main} illustrates four comparisons. The display
and Plan Cabaret cases require records of the named certification tier and
launch event; the Cassini poster requires the exhibition schedule. For the
museum, the evidence must distinguish an opening from the documented
construction stage. Each case ties the checked evidence to the depicted text,
subject, or activity.

\section{Discussion and Conclusion}

We formulated \task, in which a system verifies the situation depicted in an
image without a supplied textual claim. We introduced \benchmark{} to evaluate
both verdict accuracy and evidence sufficiency. The benchmark combines
controlled construction with web collection to cover designed and circulating
images. We developed \ours{} to refine
factual questions through visual inspection and external evidence gathering.
We trained a compact policy using full-trajectory SFT followed by PSD on
verified decisions from its own investigations.\looseness=-1

Across five shared backbones, \ours{} improves evidence sufficiency over direct
judgment. The trained Qwen3.5-9B policy achieves 79.05\% balanced accuracy and
52.26\% SESR, the highest SESR among the evaluated agents. Removing web search,
image retrieval, or evidence inspection reduces both metrics. These findings
support verifying image content through evidence about the depicted situation,
rather than judging factuality from visual plausibility or image origin.

\Needspace{10\baselineskip}
\subsection*{AI use statement}

Generative AI tools were used to create synthetic images for the controlled
construction branches of \benchmark{}. Each generated example followed a
predefined factual situation or factual-error type grounded in the associated
construction record. The authors manually reviewed the generated candidates
for visual clarity and consistency with the intended factual content, and take
responsibility for the final dataset and all results reported in this paper.
We also used generative AI tools to polish the manuscript's language. All such
revisions were reviewed and finalized by the authors.

\setlength{\bibsep}{0pt}
\bibliography{references}
\bibliographystyle{plainnat}

\clearpage
\appendix

\section{\benchmark{} Construction and Annotation}
\label{app:construction}

This appendix details the construction, factual annotation, error types,
expert review, and composition of \benchmark{}.

\subsection{Relation to Existing Benchmarks}
\label{app:benchmark-comparison}

Table~\ref{tab:benchmark-comparison} locates \benchmark{} among adjacent
resources for generated-image detection, out-of-context detection, and
multimodal fact-checking. ``No claim'' means that no claim, caption, post, or
generation prompt is supplied; ``World fact'' requires a verdict about the
depicted situation rather than provenance, plausibility, or pairwise matching.
``Evidence output'' requires evidence-bearing justification, not a label or
free-form rationale alone. Origin-independent factuality means that image
origin alone does not determine the factual label or score.

\begin{table}[!htbp]
\centering
\caption{Comparison with representative benchmarks.}
\label{tab:benchmark-comparison}
\scriptsize
\setlength{\tabcolsep}{3pt}
\renewcommand{\arraystretch}{1.18}
\begin{tabular}{@{}>{\raggedright\arraybackslash}p{0.25\linewidth}
                  *{5}{>{\centering\arraybackslash}p{0.095\linewidth}}
                  >{\raggedright\arraybackslash}p{0.145\linewidth}@{}}
\toprule
\textbf{Benchmark} & \textbf{No claim} & \textbf{World fact} &
\textbf{Web evidence} & \textbf{Evidence output} &
\textbf{Origin-independent factuality} & \textbf{Size} \\
\midrule
\multicolumn{7}{@{}l}{\textit{Generated-image detection and visual reasoning}} \\
GenImage \citeyearpar{NEURIPS2023_f4d4a021} & \yesmark & \nomark & \nomark & \nomark & \nomark & 2.68M images \\
WHOOPS! \citeyearpar{10377616} & \yesmark & \nomark & \nomark & \nomark & \nomark & 500 images \\
T2I-FactualBench \citeyearpar{huang-etal-2025-t2i} & \nomark & \yesmark & \nomark & \nomark & \yesmark & 3K prompts \\
\midrule
\multicolumn{7}{@{}l}{\textit{Image-text and out-of-context verification}} \\
NewsCLIPpings \citeyearpar{luo-etal-2021-newsclippings} & \nomark & \nomark & \nomark & \nomark & \nomark & 988,283 pairs \\
COSMOS \citeyearpar{aneja2023cosmos} & \nomark & \nomark & \nomark & \nomark & \nomark & 204,458 images \\
VERITE \citeyearpar{papadopoulos2024verite} & \nomark & \nomark & \nomark & \nomark & \yesmark & 1K pairs \\
\midrule
\multicolumn{7}{@{}l}{\textit{Multimodal fact-checking}} \\
MOCHEG \citeyearpar{10.1145/3539618.3591879} & \nomark & \yesmark & \yesmark & \yesmark & \nomark & 15,601 claims \\
FACTIFY3M \citeyearpar{chakraborty2023factify3mbenchmarkmultimodalfact} & \nomark & \yesmark & \yesmark & \yesmark & \yesmark & 3M samples \\
AVerImaTeC \citeyearpar{NEURIPS2025_0695cbd1} & \nomark & \yesmark & \yesmark & \yesmark & \nomark & 1,297 claims \\
VeriTaS \citeyearpar{rothermel-etal-2026-veritas} & \nomark & \yesmark & \yesmark & \yesmark & \yesmark & 25K claims \\
RW-Post \citeyearpar{xu2026rwpostauditableevidencegroundedmultimodal} & \nomark & \yesmark & \yesmark & \yesmark & \nomark & 1.52K + 218 cases \\
\midrule
\rowcolor{resultshade}
\textbf{\benchmark{} (ours)} & \yesmark & \yesmark & \yesmark & \yesmark & \yesmark & \textbf{10,174 images} \\
\bottomrule
\end{tabular}
\end{table}

\benchmark{} combines image-based verification without a supplied textual
claim, external reference evidence, and origin-independent factual labels.
Generated images may be supported or refuted according to the factual content
they depict. Evaluation considers both verdict accuracy and the sufficiency
of the evidence stated in the explanation.

\subsection{Construction and Collection}
\label{app:dataset-acquisition}

The curation process combines controlled construction and web collection,
as shown in Figure~\ref{fig:benchmark-pipeline}. For controlled construction,
the agent first establishes a factual situation from reference material and
specifies the content to render. Supported images express that situation;
refuted images introduce a selected factual error. For web collection, the
agent acquires an image, recovers the source information needed to interpret
it, and assigns its factual label and, where applicable, error type.
Table~\ref{tab:acquisition-pathways} summarizes the five acquisition pathways.

\begin{table}[!htbp]
\caption{Acquisition pathways, retained examples, and the basis for their factual interpretation.}
\label{tab:acquisition-pathways}
\label{tab:route-composition}
\centering
\small
\setlength{\tabcolsep}{4pt}
\renewcommand{\arraystretch}{1.20}
\begin{tabular}{@{}>{\raggedright\arraybackslash}p{0.245\linewidth}
                  >{\raggedright\arraybackslash}p{0.105\linewidth}
                  >{\raggedleft\arraybackslash}p{0.065\linewidth}
                  >{\raggedright\arraybackslash}p{\dimexpr0.585\linewidth-24pt\relax}@{}}
\toprule
\textbf{Pathway} & \textbf{Label} & \textbf{Cases} & \textbf{Factual basis} \\
\midrule
Synthesis from factual references & Supported & 2,073 &
Reference material establishes the situation that the generated image is intended to depict. \\
\addlinespace[2pt]
Supported web images & Supported & 1,041 &
Source information corroborates the depicted situation. \\
\addlinespace[2pt]
Counterfactual construction & Refuted & 3,106 &
A documented situation is modified through a selected visible detail or relation. \\
\addlinespace[2pt]
Fabrication from real entities & Refuted & 3,067 &
A constructed occurrence or relation conflicts with records or constraints applying to the depicted subjects. \\
\addlinespace[2pt]
Images from fact-checking sources & Refuted & 887 &
A circulating image is paired with source material explaining its factual error. \\
\midrule
\multicolumn{2}{@{}l}{\textbf{Total}} & \textbf{10,174} &
3,114 supported and 7,060 refuted images. \\
\bottomrule
\end{tabular}
\end{table}

In counterfactual construction, the source situation supplies the context in
which a chosen detail or relation is changed. Fabrication instead composes an
occurrence involving recognizable entities and checks it against the relevant
records or constraints. Both pathways use the taxonomy to guide construction
and then inspect the resulting image for the intended discrepancy. Collected
cases retain the factual discrepancy documented by their source material.
Collection includes both photographs and synthetic images that have circulated
online.

\subsection{Factual Annotation and Evidence}
\label{app:factual-annotation}

The annotation connects the image to the factual situation that determines its
interpretation. It describes the relevant subjects, activity or relation, and
context, together with the evidence supporting or refuting that situation.
Construction records preserve the intended factual content and any selected
change. Collection records preserve the source information used to identify
and interpret the visual material. They also provide private evaluation
references.\looseness=-1

For generated supported images, review concerns the identities, activities,
relationships, and contextual details expressed in the completed image.
Reference material establishes the factual basis, and the rendered content
is checked against it. For refuted images, the explanation identifies the depicted
relation that conflicts with the associated evidence. Visual references assess
appearance and composition, while factual interpretation follows the associated
evidence.\looseness=-1

\subsection{Error-Type Definitions}
\label{app:taxonomy-definitions}

Table~\ref{tab:taxonomy-definitions-full} expands the twelve types shown in
Figure~\ref{fig:benchmark-taxonomy}. For taxonomy-driven controlled
construction, each refuted image is assigned the type that specifies its
intended factual discrepancy. Collected cases use the factual error documented
by their source records as the annotation.

\begin{table}[!htbp]
\caption{Definitions of the twelve factual error types in \benchmark{}.}
\label{tab:taxonomy-definitions-full}
\centering
\small
\setlength{\tabcolsep}{5pt}
\renewcommand{\arraystretch}{1.35}
\begin{tabular}{@{}>{\raggedright\arraybackslash}p{0.25\linewidth}
                  >{\raggedright\arraybackslash}p{\dimexpr0.75\linewidth-10pt\relax}@{}}
\toprule
\textbf{Error type} & \textbf{Definition} \\
\midrule
\multicolumn{2}{@{}l}{\textbf{Event and relation errors}} \\
Invented action & A real subject performs an action or participates in an event that did not occur. \\
\addlinespace[2pt]
Invented interaction & Real subjects are depicted meeting, collaborating, or participating together when that interaction did not occur. \\
\addlinespace[2pt]
Impossible state & The depicted state contradicts physics, science, technology, or commonsense. \\
\addlinespace[2pt]
False official record & A visible roster, certificate, scoreboard, directory, or similar display presents an incorrect official role, result, ranking, or entry. \\
\midrule
\multicolumn{2}{@{}l}{\textbf{Attribute and detail errors}} \\
Numerical alteration & A visible number, count, score, measurement, or statistic has an incorrect value. \\
\addlinespace[2pt]
Identity substitution & The depicted person, institution, product, affiliation, or role is misidentified. \\
\addlinespace[2pt]
Spatial alteration & An object's placement, orientation, relative position, or local relation is wrong. \\
\addlinespace[2pt]
Text or mark alteration & Visible wording, a logo, flag, formula, or short label contains incorrect content. \\
\midrule
\multicolumn{2}{@{}l}{\textbf{Contextual errors}} \\
Temporal displacement & Visible content places a subject or event at the wrong date, year, era, or time. \\
\addlinespace[2pt]
Event recontextualization & An identifiable scene or event is presented as a different occurrence through the image or its visible presentation. \\
\addlinespace[2pt]
Source impersonation & A visible post, page, account, publisher, verification marker, or platform identity is falsely attributed. \\
\addlinespace[2pt]
Geographic displacement & The stated venue, landmark, city, country, or region is wrong. \\
\bottomrule
\end{tabular}
\end{table}

The primary discrepancy also resolves overlap between neighboring types.
Identity substitution concerns who or what occupies a role; false official
record concerns the association or result asserted by a record. A changed
number is a numerical alteration when its value is the central error, while
an incorrect official outcome is assigned to false official record. Spatial
alteration concerns local arrangement, and geographic displacement concerns
the location associated with the scene. Temporal displacement changes the
time; event recontextualization changes the represented occurrence.\looseness=-1

\subsection{Expert Review}
\label{app:dataset-screening}

Expert review follows automated construction and collection. Two domain experts
review each image together with its factual description and retained records.
They check the clarity of decisive visual details and text, verify that the
target is grounded in visible content, and assess whether the evidence supports
the required identities and context and justifies the assigned label. For
generated cases, they also check that the intended fact or discrepancy appears
in the image; for refuted cases, they confirm the primary error type. A third
expert resolves disagreements.

Review decisions are acceptance, acceptance with edits, or rejection. Edits
address factual wording, source association, and category assignment. Unclear
or unresolved cases are rejected.

\subsection{Composition, Splits, and Availability}
\label{app:dataset-splits}

The dataset has 10,174 images across five pathways
(Table~\ref{tab:acquisition-pathways}), with 8,647 in training and 1,527 in
the test set. Table~\ref{tab:source-split-composition} gives the source and
factual-label counts for both partitions. All retained images underwent expert
review; admitted investigation trajectories are reported in
Appendix~\ref{app:sft-details}.

The controlled test subset contains 513 Nano Banana Pro
\citep{google2025nanobananapro}, 423 GPT Image~2
\citep{openai2026gptimage2}, and 302 Seedream~5.0 Lite
\citep{bytedance2026seedream50lite} images. Controlled training images use Nano
Banana Pro; web-collected images retain their original sources.

Related images from the same source or event are grouped when constructing the
partitions, and the test set is used for evaluation. Systems receive only the
image and generic verification instruction; the associated annotations support
assessment. Upon acceptance, we will open-source all artifacts from this work: the complete dataset,
including images, labels, explanations, evidence and provenance records,
and fixed partitions; all data-construction, training, and evaluation code;
prompts and training trajectories; and all trained model checkpoints.

\begin{table}[!htbp]
\caption{Dataset composition by acquisition source, split, and factual label.}
\label{tab:source-split-composition}
\centering
\small
\setlength{\tabcolsep}{4pt}
\renewcommand{\arraystretch}{1.14}
\begin{tabular}{@{}lrrrrr@{}}
\toprule
& \multicolumn{2}{c}{\textbf{Training}} &
  \multicolumn{2}{c}{\textbf{Test}} & \textbf{Total} \\
\cmidrule(lr){2-3}\cmidrule(lr){4-5}
\textbf{Source} & \textbf{Supported} & \textbf{Refuted} &
\textbf{Supported} & \textbf{Refuted} & \\
\midrule
Controlled construction & 1,840 & 5,168 & 233 & 1,005 & 8,246 \\
Web collection & 897 & 742 & 144 & 145 & 1,928 \\
\midrule
\textbf{Total} & \textbf{2,737} & \textbf{5,910} &
\textbf{377} & \textbf{1,150} & \textbf{10,174} \\
\bottomrule
\end{tabular}
\end{table}

Both acquisition routes contribute supported and refuted cases to the training
and test partitions. Controlled construction supplies most refuted cases,
whereas web collection has a more even label split. The twelve intended error
types each account for 502 to 524 controlled refuted images.

\begin{table}[!htbp]
\caption{Error-type distribution of 6,173 refuted images from controlled construction.}
\label{tab:error-type-composition}
\centering
\small
\setlength{\tabcolsep}{5pt}
\renewcommand{\arraystretch}{1.12}
\begin{tabular}{@{}lrlr@{}}
\toprule
\textbf{Error type} & \textbf{Cases} & \textbf{Error type} & \textbf{Cases} \\
\midrule
Invented action & 513 & Numerical alteration & 520 \\
Invented interaction & 512 & Identity substitution & 504 \\
Impossible state & 502 & Spatial alteration & 514 \\
False official record & 516 & Text or mark alteration & 521 \\
Temporal displacement & 515 & Source impersonation & 510 \\
Event recontextualization & 522 & Geographic displacement & 524 \\
\bottomrule
\end{tabular}
\end{table}

\section{Implementation and Training Details}
\label{app:implementation}

This section records the configuration, runtime, and training procedures for
the reported Qwen3.5-9B policy. The prompt templates appear in
Appendix~\ref{app:method-prompts}.

\subsection{Model and Training Configuration}
\label{app:complete-configuration}

\paragraph{Compared model sources.}
The evaluated models follow the official releases or model cards for GPT-5.5
\citep{openai2026gpt55}, GPT-5.2 \citep{openai2025gpt52}, Gemini 3.7 Flash
\citep{google2026gemini37flash}, Gemini 3.1 Pro
\citep{googledeepmind2026gemini31pro}, Claude Opus 4.8
\citep{anthropic2026opus48}, Kimi K2.6 \citep{moonshot2026kimik26}, and
Doubao Seed 2.0 Pro \citep{bytedance2026seed20}. The remaining sources are
MiniMax M3 \citep{minimax2026m3}, GLM-4.6V \citep{zai2025glm46v},
Qwen3.5-397B-A17B and Qwen3.5-9B \citep{qwen3.5}.

\begingroup
\hbadness=10000
\small
\setlength{\LTleft}{0pt}
\setlength{\LTright}{0pt}
\setlength{\tabcolsep}{6pt}
\renewcommand{\arraystretch}{1.16}
\begin{longtable}{@{}>{\raggedright\arraybackslash}p{0.32\linewidth}
                     >{\raggedright\arraybackslash}p{\dimexpr0.68\linewidth-12pt\relax}@{}}
\caption{Implementation and training settings for the Qwen3.5-9B full policy.}
\label{tab:full-implementation-settings}\\
\toprule
\textbf{Hyperparameter} & \textbf{Value} \\
\midrule
\endfirsthead
\multicolumn{2}{@{}l}{\small\itshape Table~\ref{tab:full-implementation-settings} continued}\\
\toprule
\textbf{Hyperparameter} & \textbf{Value} \\
\midrule
\endhead
\midrule
\multicolumn{2}{r@{}}{\small Continued on next page}\\
\endfoot
\bottomrule
\endlastfoot

\multicolumn{2}{@{}c@{}}{\itshape Model and supervision} \\
\midrule
Policy backbone & Qwen3.5-9B \citep{qwen3.5} \\
Policy parameters & 9,409,813,744 \\
SFT demonstration model & Qwen3.5-397B-A17B \citep{qwen3.5} \\
Model-mediated tool calls & Same checkpoint as the investigation policy; separate requests \\
OCR service & Baidu general OCR API \\
SFT training set & 4,872 reviewed complete investigations \\
SFT update & Full-parameter update of language, vision, and multimodal connector modules \\
SFT schedule & 3 epochs; 3,084 optimizer steps \\
Supervised outputs & Assistant reasoning, native tool calls, and terminal reports \\
Conditioning content & Instructions, raw tool observations, and images available at each turn \\
Token loss weights & \shortstack[l]{Reasoning: 1; tool calls and report: 2;\\ instructions and tool observations: 0} \\
\midrule
\multicolumn{2}{@{}c@{}}{\itshape Sequence and image processing} \\
\midrule
Maximum context length & 131,072 tokens \\
Packing & Enabled; 120,000-token packing length; grouped by length \\
Truncation & Disabled; over-length examples raise an error \\
Image token limit & 1,024 tokens per image \\
Image pixel limit & 262,144 pixels \\
Image placement & Original image at the first turn; retrieved and derived images at their observation boundary \\
Padding-free training & Enabled \\
\midrule
\multicolumn{2}{@{}c@{}}{\itshape SFT optimization} \\*
Learning rate & $1\times10^{-5}$ \\*
Microbatch and accumulation & 1 sample; 1 accumulation step \\*
Numerical precision & BF16 \\
Attention implementation & FlashAttention 2.8.3 \\
Distributed training & FSDP2 with sequence parallel size 4 \\
Gradient checkpointing & Disabled \\
Kernel optimization & Liger 0.8.0 \\
Checkpoint interval & 400 optimizer steps; three recent checkpoints retained \\
\midrule
\multicolumn{2}{@{}c@{}}{\itshape PSD optimization} \\*
Initialization & Three-epoch SFT checkpoint at step 3,084 \\
PSD frozen self-teacher & Frozen copy of the current Qwen policy at the start of the round \\
PSD intervention model & Gemini 3.7 Flash \citep{google2026gemini37flash} for decision localization, hint construction, and continuation review \\
Student context & Original recorded policy context \\
Distillation target & Frozen-policy top-20 token distribution on accepted repair and preservation decisions \\
Adapter & LoRA rank 32; $\alpha=32$; dropout 0; all linear layers \\
Training schedule & 5 epochs \\
Learning rate & $4\times10^{-5}$ \\
Optimizer & AdamW; $\beta_1=0.9$, $\beta_2=0.95$, $\epsilon=10^{-12}$, weight decay 0 \\
Scheduler & Constant; no warmup \\
Microbatch and accumulation & 1 target; 32 accumulation steps \\
Maximum gradient norm & 1.0 \\
Loss computation & 128-token chunks over supervised completion positions \\
Numerical precision & BF16 \\
Distributed training & FSDP2 with sequence parallel size 4 \\
\midrule
\multicolumn{2}{@{}c@{}}{\itshape Inference} \\*
Policy prompts & \texttt{unified-react-raw-history-loop-v14-en}; \texttt{unified-react-raw-history-judgment-v6-en} \\
Maximum actions & 24; early completion through \texttt{finish\_investigation} \\
Sampling & Temperature 1.0; top-$p$ 0.95; top-$k$ 20; min-$p$ 0 \\
Penalties & Presence 1.5; repetition 1.0 \\
Maximum output & 32,768 tokens per model request \\
Reasoning budget & 8,192 tokens per model request \\
Retained context & Original image and complete retained tool interaction \\
\midrule
\multicolumn{2}{@{}c@{}}{\itshape Hardware and software} \\*
Hardware & 4 $\times$ NVIDIA H20 GPUs; 97,871 MiB memory per GPU \\
Training framework & ms-swift 4.4.2 \citep{zhao2025swiftascalablelightweightinfrastructure} \\
Core software & Python 3.12; PyTorch 2.10.0; CUDA 12.8; Transformers 5.12.1 \\
\end{longtable}
\endgroup

\subsection{Agent Runtime and Tools}
\label{app:agent-configuration}

Each investigation turn produces one native tool call. The runtime validates its
arguments and remaining budget, executes it, and appends the raw result and any
associated images to the history. The policy uses that history in the next
turn. Web search uses the Serper API \citep{serper2026api}. Fact-checking
websites are blocked from the agent's evidence access, including their
exclusion from web-search queries. Page inspection uses Jina Reader
\citep{jina2026reader} to fetch webpage content, followed by model-based
extraction of passages relevant to the active question. For Qwen3.5-9B, the
investigation policy and model-mediated tools use the same checkpoint within
each configuration: Base, SFT, or Full. Model-mediated visual inspection and
page-evidence extraction are executed through separate requests. Positioned
text is obtained through the Baidu general OCR API \citep{baidu2026ocr}. Search and image retrieval return candidates;
page reading or image comparison checks them before citation.

\begin{table}[!htbp]
\centering
\caption{Native operations available to the investigation policy. Numeric limits
are per case and remain subject to the 24-action global budget. ``Global'' means
that only the global budget applies.}
\label{tab:agent-tools}
\fontsize{8}{9}\selectfont
\renewcommand{\arraystretch}{1.13}
\setlength{\tabcolsep}{3.5pt}
\begin{tabular}{@{}>{\raggedright\arraybackslash}p{0.32\linewidth}
                  >{\raggedright\arraybackslash}p{\dimexpr0.56\linewidth-14pt\relax}
                  >{\centering\arraybackslash}p{0.12\linewidth}@{}}
\toprule
\textbf{Operation} & \textbf{Role and returned information} & \textbf{Limit} \\
\midrule
\texttt{perceive\_scene} & Scene description, entities, attributes, relations, and visible uncertainties. & Global \\
\texttt{ocr\_with\_position} & Text transcription with positions for later cropping and inspection. & 3 \\
\texttt{crop\_and\_inspect} & Deterministic crop and focused description of a selected region. & 4 \\
\texttt{focused\_visual\_inspection} & Original image and derived views assessed against one evidence-motivated visual question. & 2 \\
\texttt{analyze\_visual\_anomalies} & Locatable physical, logical, relation, or text observations relevant to the current target. & 3 \\
\texttt{check\_consistency} & Lighting, geometry, scale, boundary, and physical-relation checks for a specified aspect. & 3 \\
\texttt{count\_objects} & Count and localization of a requested object type. & Global \\
\texttt{text\_search} & Ranked web results for a concrete factual query. & 16 \\
\texttt{reverse\_image\_search} & Visually related images and their source pages from the image or a crop. & 2 \\
\texttt{text\_image\_search} & Candidate reference images and associated pages from a textual query. & 6 \\
\texttt{visit} & Relevant passages and source metadata from one to three selected URLs. & 16 \\
\texttt{compare\_with\_reference} & Same-subject or same-event relation and concrete differences between the input and a selected reference. & 6 \\
\texttt{current\_time} & Current date and time when temporal interpretation requires it. & 1 \\
\bottomrule
\end{tabular}
\end{table}

\paragraph{Ablation configuration.}
Each tool ablation disables one of the three tool families listed below, while
all remaining tools stay available. Full uses the complete tool set. All
variants use the Full policy checkpoint and share the prompt, action budget,
and terminal-judgment procedure.

\begin{table}[!htbp]
\centering
\caption{Native-operation settings for the tool ablations in Table~\ref{tab:ablations}.}
\label{tab:tool-ablation-configuration}
\small
\setlength{\tabcolsep}{6pt}
\renewcommand{\arraystretch}{1.12}
\begin{tabular}{@{}>{\raggedright\arraybackslash}p{0.32\linewidth}
                  >{\raggedright\arraybackslash}p{\dimexpr0.68\linewidth-12pt\relax}@{}}
\toprule
\textbf{Variant} & \textbf{Disabled native operations} \\
\midrule
Full & None \\
w/o web search & \texttt{text\_search} \\
w/o image retrieval & \texttt{text\_image\_search}, \texttt{reverse\_image\_search} \\
w/o evidence inspection & \texttt{visit}, \texttt{compare\_with\_reference} \\
\bottomrule
\end{tabular}
\end{table}

The policy may finish before the limit through
\texttt{finish\_investigation}. Completion or exhaustion of the action budget
passes the retained interaction to terminal judgment. This stage receives the
original image and accumulated observations, then produces the verdict and
report. Citations identify successful observations used in the explanation.
Empty results and failed requests remain in the history with their original
status.

Web and text-to-image search use the Serper web and image endpoints. Image
search returns at most three candidate image and page pairs. Reverse-image
retrieval uses Serper Lens; its semantic branch generates one visual query and
submits it to image search. Page inspection fetches one to three selected URLs
and extracts passages for the current image fact and retrieval goal. Visual
tools receive the original image or a deterministic crop and return structured
observations.

\subsection{SFT Data and Supervision}
\label{app:sft-details}

Qwen3.5-397B-A17B \citep{qwen3.5} generates the teacher
investigations through the same tool interface used at inference time. Each
retained example includes the complete action and observation sequence, the
images available at successive turns, and a terminal report supported by the
recorded investigation. For each training image, the teacher agent performs up
to three rollout attempts. Once a complete trajectory passes review, it is
retained and no further attempt is made for that image. We perform
full-parameter fine-tuning of Qwen3.5-9B
\citep{qwen3.5} on the 4,872 reviewed investigations. The weighted
supervision emphasizes native tool calls and the terminal report while retaining
the reasoning that led to them.

\subsection{PSD Procedure}
\label{app:psd-details}

Table~\ref{tab:psd-procedure} follows one PSD round from fresh investigations to
accepted training targets and the policy update. Gemini 3.7 Flash
\citep{google2026gemini37flash} proposes and reviews the intervention; the
frozen Qwen policy supplies the continuation and token distribution.

\begin{table}[!htbp]
\centering
\caption{One PSD round. Symbols follow the formulation in the main text.}
\label{tab:psd-procedure}
\small
\renewcommand{\arraystretch}{1.22}
\begin{tabular}{@{}r p{0.91\linewidth}@{}}
\toprule
1 & Freeze the current policy as $\pi_{\bar\theta}$ and collect fresh investigations. \\
2 & Select a decision $t$ in an eligible student investigation. \\
3 & Construct a procedural hint $h_t$ for that decision. \\
4 & Continue $\pi_{\bar\theta}$ from $\operatorname{insert}(c_t,h_t)$ through terminal judgment. \\
5 & Review the selected decision, completed investigation, evidence use, and trajectory structure. \\
6 & Add an accepted completion $u_d$ with teacher context $c_d^{\mathrm T}=\operatorname{insert}(c_t,h_t)$ and student context $c_d^{\mathrm S}=c_t$ to $\mathcal D_{\mathrm{rep}}$. \\
7 & Extract preservation decisions from successful investigations with $c_d^{\mathrm T}=c_d^{\mathrm S}$ into $\mathcal D_{\mathrm{pres}}$. \\
8 & Store the frozen policy's top-$K$ probabilities for each supervised completion token. \\
9 & Update $\pi_\theta$ on $\mathcal D_{\mathrm{rep}}\cup\mathcal D_{\mathrm{pres}}$ using $\mathcal L_{\mathrm{PSD}}$. \\
10 & Use the updated policy to begin the next rollout round. \\
\bottomrule
\end{tabular}
\end{table}

For each target, the teacher and student receive the same recorded image inputs.
Repair targets pair a hint-augmented teacher context with the original student
context. The distillation loss covers the selected assistant completion, using
renormalized top-$K$ probabilities from the frozen teacher. Successful
investigations supply preservation targets with identical teacher and student
contexts. Repair and preservation targets use the same row weight.\looseness=-1

For target \(d\) and completion position \(\ell\), the frozen teacher defines
\[
q_{d,\ell}(v)
=p_{\bar\theta}\!\left(v\mid c_d^{\mathrm T},u_{d,<\ell}\right),
\qquad
\mathcal K_{d,\ell}=\operatorname{TopK}_{v}\,q_{d,\ell}(v),
\]
and the retained probabilities are renormalized as
\[
\bar q_{d,\ell}(v)
=
\frac{q_{d,\ell}(v)}
{\sum_{v'\in\mathcal K_{d,\ell}}q_{d,\ell}(v')},
\qquad v\in\mathcal K_{d,\ell}.
\]

\section{Evaluation Protocol}
\label{app:evaluation}

\subsection{Inputs and Evaluation Settings}

Table~\ref{tab:eval-settings} collects the common test input, evaluator, and
execution settings. The test images are separate from teacher trajectories and
SFT or PSD training data.

\begin{table}[!htbp]
\centering
\caption{Test set and evidence evaluation settings.}
\label{tab:eval-settings}
\small
\setlength{\tabcolsep}{6pt}
\begin{tabular}{@{}>{\raggedright\arraybackslash}p{0.32\linewidth}
                  >{\raggedright\arraybackslash}p{\dimexpr0.68\linewidth-12pt\relax}@{}}
\toprule
\textbf{Setting} & \textbf{Value} \\
\midrule
Test images & 1,527: 377 supported and 1,150 refuted \\
Evaluator & Gemini 3.7 Flash (\texttt{gemini-3.7-flash}) \citep{google2026gemini37flash} \\
Classification statistics & BAcc, macro F1, and class recalls \\
Evidence evaluation & Stated grounds checked against private reference material \\
Evaluation population & All 1,527 test images \\
Execution attempts & Up to three per case; stop at the first binary verdict \\
\bottomrule
\end{tabular}
\end{table}

\subsection{Metrics}

For each class $c\in\mathcal{C}=\{\mathrm{s},\mathrm{r}\}$, let $TP_c$,
$FP_c$, and $FN_c$ denote true positives, false positives, and false negatives.
The class supports are $n_{\mathrm{s}}=377$ and $n_{\mathrm{r}}=1150$.
We compute recall and F1 as
\begin{equation}
R_c=\frac{TP_c}{n_c},\qquad
F1_c=\frac{2TP_c}{2TP_c+FP_c+FN_c}.
\end{equation}
Both classes receive equal weight in BAcc and macro F1:
\begin{equation}
\mathrm{BAcc}=\frac{1}{2}\sum_{c\in\mathcal{C}}R_c,\qquad
\mathrm{MacroF1}=\frac{1}{2}\sum_{c\in\mathcal{C}}F1_c.
\label{eq:appendix-bacc}
\end{equation}

Let $e_i=1$ when the verdict is correct and the response itself provides
sufficient grounds for the decisive image-grounded fact; otherwise, $e_i=0$.
We compute SESR over all 1,527 images as
\begin{equation}
\mathrm{SESR}=\frac{1}{1527}\sum_{i=1}^{1527}e_i.
\label{eq:appendix-sesr}
\end{equation}
All metrics are reported as percentages.

\paragraph{Execution attempts.}
Each case permits up to three attempts, stopping at the first response with the
required binary verdict. Cases with content-based refusals in all three
attempts are counted as incorrect predictions for their ground-truth class.

\subsection{Assessment of Complete Responses}

The evaluator checks verdict correctness, factual alignment, and evidence
sufficiency against private reference material. The grounds stated in the
response must establish the decisive image-grounded fact or another
image-grounded sub-fact that logically establishes the same verdict. Equivalent
wording, alternative evidence sources, and different investigation paths are
accepted when these criteria are met. A response receives SESR credit when all
three criteria are satisfied.

\section{Additional Results}
\label{app:additional-experiments}

\subsection{Additional Quantitative Analysis}

\begin{figure}[!ht]
    \centering
    \includegraphics[width=\textwidth]{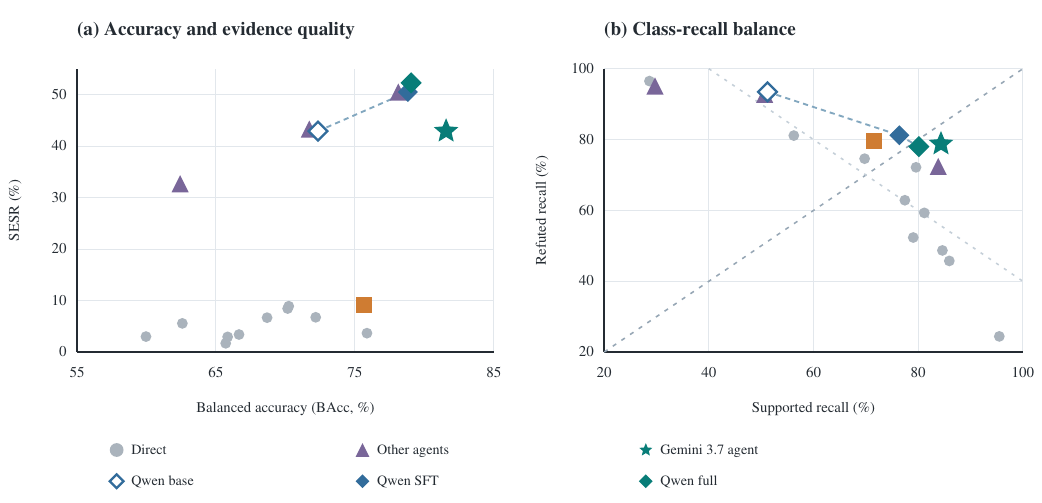}
    \caption{Performance on the test set. (a) BAcc and SESR for
    direct models and agent policies. (b) Supported and refuted recall, with
    equal-recall and equal-BAcc guides. The Qwen3.5-9B comparison includes
    the base agent, three-epoch SFT, and full SFT+PSD policies.}
    \label{fig:experiment-overview}
\end{figure}

Figure~\ref{fig:experiment-overview} presents two views of the results in
Table~\ref{tab:main-results}. Panel (a) relates verdict accuracy to
evidence sufficiency. Panel (b) separates the recalls of supported and refuted
images to show how their contributions vary across systems.

\paragraph{Accuracy and evidence sufficiency.}
Investigation consistently moves systems upward along the SESR axis, while its
effect on BAcc varies across backbones. Training moves the compact Qwen policy
toward the upper-right region with stronger verdict accuracy and greater
evidence sufficiency.

\paragraph{Recall across factual labels.}
Several direct models lie far from the equal-recall guide. For the Qwen3.5-9B
agent, SFT raises supported recall and lowers refuted recall, closing their gap.

\begin{figure}[!ht]
    \centering
    \includegraphics[width=\textwidth]{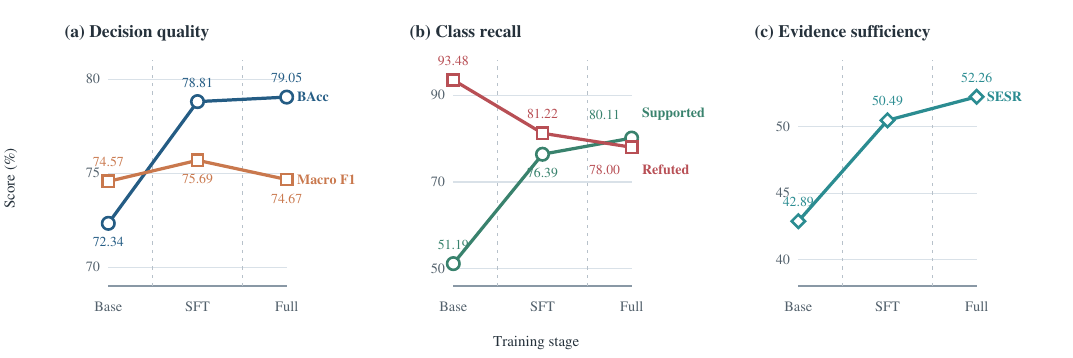}
    \caption{Development of the Qwen3.5-9B policy across training stages.
    Panels separate overall decision quality, recall for the two factual
    labels, and evidence sufficiency. Full denotes the policy trained with SFT
    and PSD. Values are percentages; vertical scales differ across panels.}
    \label{fig:training-progression}
\end{figure}

Figure~\ref{fig:training-progression} isolates changes under the same backbone.
SFT provides the largest change from the base policy. PSD further improves BAcc
and SESR while narrowing the gap between class recalls.

\FloatBarrier
\section{Prompt Templates}
\label{app:method-prompts}

This appendix presents selected prompts used for investigation, judgment, tool
use, and training. The implementation labels \texttt{real} and \texttt{fake}
correspond to \textsc{supported} and \textsc{refuted}, respectively.

\subsection{Active Policy Prompts}

The investigation prompt asks the policy to organize its notes as
\texttt{Established}, \texttt{Open}, and \texttt{Action}. The runtime validates
and executes native tool calls, then appends their results to the interaction
history.

\promptfile{EVISCENE INVESTIGATION POLICY}{prompts/eviscene-investigation-policy.txt}
\promptfile{EVISCENE TERMINAL JUDGMENT}{prompts/eviscene-terminal-judgment.txt}

\subsection{Direct-Judgment Prompt}

\promptfile{DIRECT IMAGE FACTUAL JUDGMENT}{prompts/evaluation-direct-judgment.txt}

\subsection{Representative Tool Prompts}

\promptfile{SCENE PERCEPTION}{prompts/tool-scene-perception.txt}
\promptfile{REFERENCE-IMAGE COMPARISON}{prompts/tool-reference-comparison.txt}

\subsection{Training-Only Review and Repair Prompts}

\promptfile{SFT TRAJECTORY REVIEW}{prompts/training-trajectory-review.txt}
\promptfile{PSD DECISION LOCALIZATION}{prompts/training-psd-localization.txt}
\paragraph{PSD hint construction.}
The following listing reproduces the fixed prompt used to construct and revise
decision-local procedural hints.
\promptfile{PSD HINT CONSTRUCTION}{prompts/training-psd-hint-construction.txt}
\promptfile{PSD REPAIR REVIEW}{prompts/training-psd-repair-review.txt}

\end{document}